\documentclass[10pt,twocolumn,letterpaper]{article}

\usepackage[pagenumbers]{cvpr} 

\usepackage[dvipsnames]{xcolor}

\definecolor{cvprblue}{rgb}{0.21,0.49,0.74}
\usepackage[pagebackref,breaklinks,colorlinks,citecolor=cvprblue]{hyperref}

\usepackage{xcolor}
\usepackage{multirow}

\usepackage[capitalize]{cleveref}
\crefname{section}{Sec.}{Secs.}
\Crefname{section}{Section}{Sections}
\Crefname{table}{Table}{Tables}
\crefname{table}{Tab.}{Tabs.}

\newcommand{\hz}{\vphantom{\parbox[c]{0.25cm}{\rule{0.25cm}{0.28cm}}}}
\newcommand{\frozen}[1]{{\setlength\fboxsep{1.5pt}\colorbox{blue!20}{\hz{$\displaystyle #1$}}}}
\newcommand{\learn}[1]{{\setlength\fboxsep{1.5pt}\colorbox{red!20}{\hz{$\displaystyle #1$}}}}

\def\paperID{*****} 
\def\confName{CVPR}
\def\confYear{2024}

\title{MARS: Mixture of Auto-Regressive Models for \\ Fine-grained Text-to-image Synthesis} 

\vspace{-1cm}
\author{
Wanggui He\textsuperscript{1,\textbf{*}},
Siming Fu\textsuperscript{1,\textbf{*}},
Mushui Liu\textsuperscript{2,\textbf{*}},
Xierui Wang\textsuperscript{2,+},
Wenyi Xiao\textsuperscript{2,+},
Fangxun Shu\textsuperscript{1,+}, \\
Yi Wang\textsuperscript{2},
Lei Zhang\textsuperscript{2},
Zhelun Yu\textsuperscript{3},
Haoyuan Li\textsuperscript{2}, 
Ziwei Huang\textsuperscript{2},
LeiLei Gan\textsuperscript{2},
Hao Jiang\textsuperscript{1,\textbf{\textdagger}},
\\
\textsuperscript{1}Alibaba Group~~~~\textsuperscript{2}Zhejiang University~~~~\textsuperscript{3}Fudan University
\\
\small \textsuperscript{\textbf{*}}Equal contribution~~~~ \small \textsuperscript{\textbf{+}}Core contributor~~~~ \small \textsuperscript{\textbf{\textdagger}}Corresponding author
}

\begin{document}

\twocolumn[{%
\renewcommand\twocolumn[1][]{#1}%
\maketitle
\begin{center}
    \centering
    \captionsetup{type=figure}
    \vspace{-5mm}
    \includegraphics[width=\linewidth]{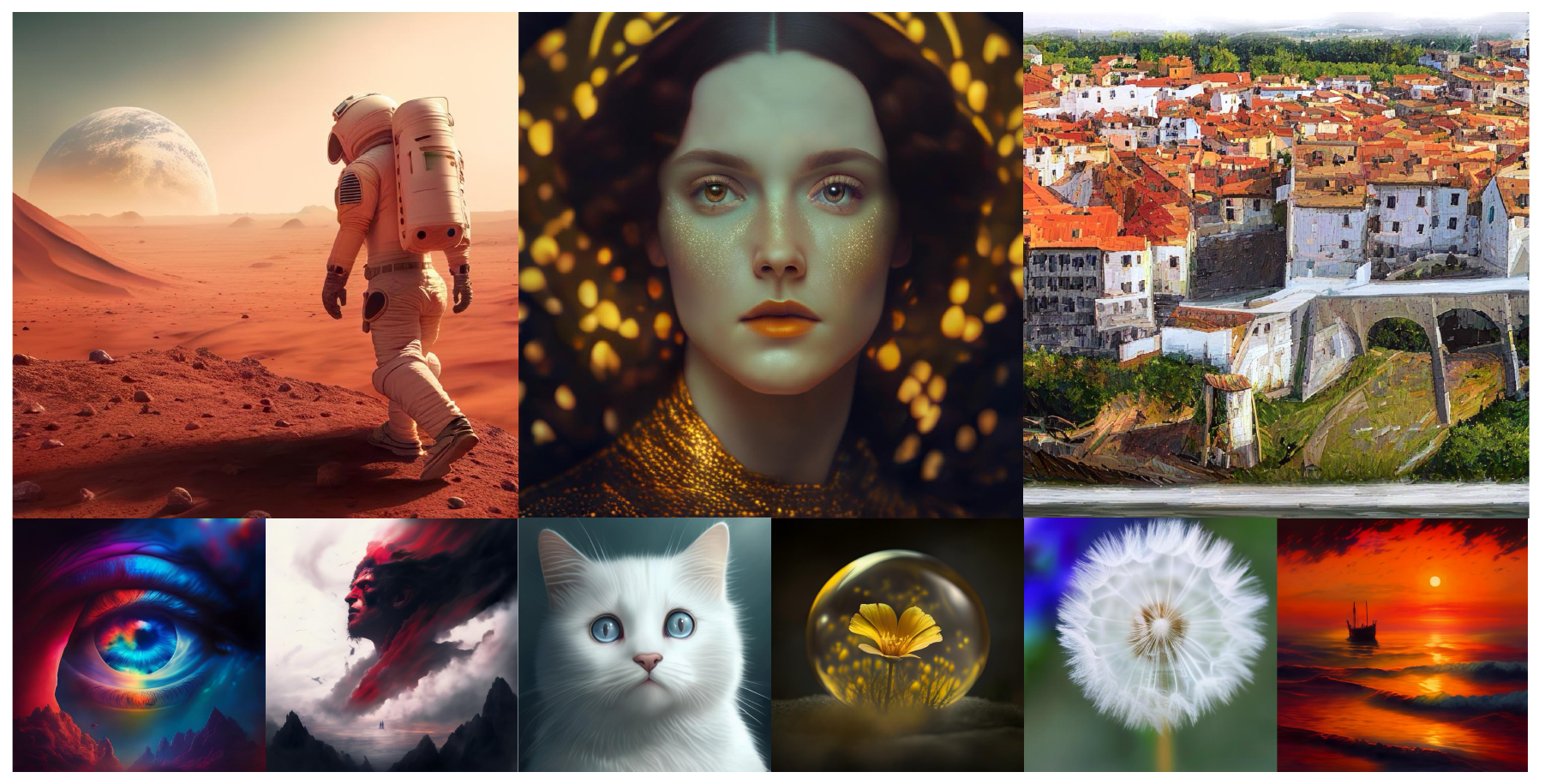}
    \vspace{-7.5mm}
    \captionof{figure}{The generated samples from MARS display extraordinary quality, marked by an impressive degree of fidelity and precision in their adherence to the provided textual descriptions. }
    \label{fig:teaser}
\end{center}%
}]

\begin{abstract}
Auto-regressive models have made significant progress in the realm of language generation, yet do not perform on par with diffusion models in the domain of image synthesis. In this work, we introduce \textbf{MARS}, a novel framework for T2I generation that incorporates a specially designed Semantic Vision-Language Integration Expert (SemVIE). This innovative component integrates pre-trained LLMs by independently processing linguistic and visual information—freezing the textual component while fine-tuning the visual component. This methodology preserves the NLP capabilities of LLMs while imbuing them with exceptional visual understanding. Building upon the powerful base of the pre-trained Qwen-7B, MARS stands out with its bilingual generative capabilities corresponding to both English and Chinese language prompts and the capacity for joint image and text generation.   The flexibility of this framework lends itself to migration towards \textbf{any-to-any} task adaptability. Furthermore, MARS employs a multi-stage training strategy that first establishes robust image-text alignment through complementary bidirectional tasks and subsequently concentrates on refining the T2I generation process, significantly augmenting text-image synchrony and the granularity of image details.  Notably, MARS requires only \textbf{9\%} of the GPU days needed by SD1.5, yet it achieves remarkable results across a variety of benchmarks, illustrating the training efficiency and the potential for swift deployment in various applications. Code will be available at \url{https://github.com/fusiming3/MARS}.
\end{abstract} 
\section{Introduction}
\label{sec:intro}

Pre-trained Large Language Models (LLMs) \cite{zhang2022opt,brown2020language,wei2021finetuned,touvron2023llama,vicuna2023} have broadened their generative capabilities to encompass the visual domain. This advancement entails transforming pixel data into discrete tokens through a visual tokenizer, analogous to the processing of textual information, thereby integrating these tokens into the model's transformer \cite{vaswani2017attention} architecture for generative tasks. Unlike other generative approaches, such as diffusion models \cite{rombach2022high, podell2023sdxl, esser2024scaling, chen2023pixart}, LLMs~\cite{Muse, yu2023scaling, cogview2, ma2024star} uniquely utilize a discrete latent space of visual tokens, crucial for merging visual and linguistic modalities. 

Auto-regressive models for text-to-image generation models, such as Parti \cite{yu2023scaling}, CogView2 \cite{cogview2}, and Unified-io2 \cite{lu2024unified} have extended their generative scope to encompass the visual domain, facilitating the creation of images. These models integrate pre-trained LLMs within a unified architecture, enabling the simultaneous interpretation of both linguistic and visual inputs. Nonetheless, a notable challenge arises from the inherent distributional bias of LLMs, which are predominantly trained on textual data, potentially leading to a pronounced distributional shift when adapting to text-image pair datasets. This shift has the potential to provoke catastrophic forgetting, consequently impairing the LLMs' primary competency in text generation tasks. \textit{The aforementioned discourse prompts a pivotal inquiry: is it feasible to preserve the natural language processing proficiency of LLM while concurrently endowing it with state-of-the-art visual comprehension and generation capabilities?}

In response to this challenge, we present MARS, an innovative framework predicated on an auto-regressive model architecture akin to that of pre-trained LLMs for text-to-image synthesis. Specifically, we design the Semantic Vision-Language Integration Expert (SemVIE) module as the centerpiece of MARS to seamlessly facilitate the frozen pre-trained LLM with the trainable visual expert, thereby endowing them with exceptional visual understanding and preserving the NLP capability of pre-trained LLMs. Moreover, SemVIE can facilitate a comprehensive and incremental interplay between the textual and visual modalities across every layer of the model, fostering deep integration that yields images closely aligned with their textual descriptors. Through rigorous training on paired image-text datasets, MARS augments the generative capabilities of LLMs to include sophisticated text-to-image translations. As demonstrated in \cref{fig:teaser}, MARS exhibits a pronounced ability to generate images with intricate visual details, such as animal fur, plant foliage, and facial features, underscoring its potent text-to-image generation proficiency. 

In the domain of data optimization, we have developed a content-rich, efficient, and fine-grained approach for dataset construction. We leverage the capabilities of CogVLM \cite{wang2023cogvlm} to generate sophisticated image descriptions that enhance text-to-image alignment. For the optimization of the model training process, we have devised a multi-stage training strategy. This regimen begin with the creation of low-resolution images and advancing toward the production of high-resolution images with detailed textual alignment. Remarkably, with a mere 587 A100 GPU days, equating to only \textbf{9\%} of the training duration required by Stable Diffusion v1.5, MARS demonstrates its superiority over existing large-scale text-to-image (T2I) models, as evidenced in \cref{fig: fid-training-time}. Our contributions can be encapsulated as follows:
\begin{figure}[!t]
    \centering
    \includegraphics[width=\linewidth]{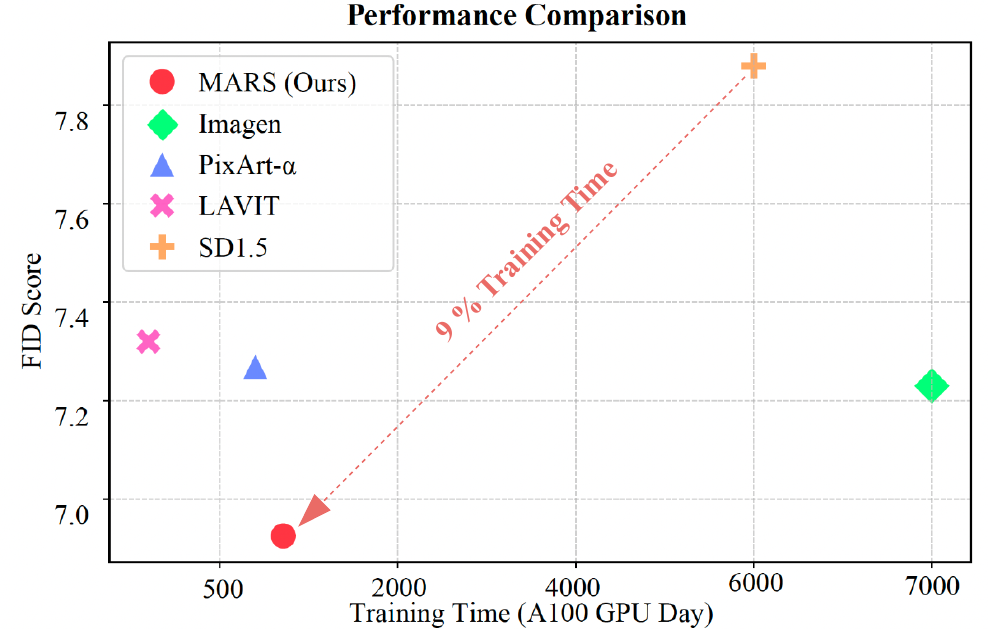}
    \caption{Comparison of training time and performance with models. The FID is evaluated on the zero-shot MS-COCO benchmark.}
    \vspace{-0.5cm}
    \label{fig: fid-training-time}
\end{figure}

\begin{itemize}
\item We present MARS, an innovative framework adapted from auto-regressive pre-trained LLMs for T2I generation tasks. To ensure preservation of NLP capacities while also equip the model with advanced visual generation and comprehension abilities, we design a module named SemVIE, which adds parallel visual experts to the attention blocks of pre-traiend LLM. Therefore, MARS amplifies the flexibility of autoregressive methods for T2I generation and joint image-text synthesis, with the potential expansibility to \textbf{any-to-any} tasks.

\item We propose a multi-stage refinement training strategy that significantly enhances MARS' robust instruction-following capability and its ability to generate high-quality images with rich details.
\item MARS shows great ability in prompt understanding and following, \textit{e.g.} long and complex nature language inputs.
Moreover, it possesses the \textbf{bilingual} capacity to follow prompts in both English and Chinese.
The framework's performance is verified across an array of evaluative measures,  \textit{i.e.} MS-COCO benchmark, T2I-CompBench, and Human Evaluation.
\end{itemize}

\section{Related Works}
\subsection{Text-to-Image Generation Models}
Text-to-image generation aims to create images based on given textual descriptions.  Recent diffusion-based models \cite{sohl2015deep,song2019generative,ho2020denoising,song2020improved,song2020score} have demonstrated exceptional performance in image generation, offering improved stability and controllability. These models operate by introducing Gaussian noise to input images in a forward process and subsequently generate high-quality images with intricate details and diversity through an inverse process starting from random Gaussian noise. Models like GLIDE \cite{nichol2021glide} and Imagen \cite{saharia2022photorealistic} utilize the CLIP \cite{radford2021learning} text encoder to enhance image-text alignment. Latent Diffusion Models (LDMs) \cite{rombach2022high} has been proposed to shift the diffusion process from pixel space to latent space, thereby enhancing efficiency and image quality. Furthermore, recent advancements such as SD-XL \cite{podell2023sdxl}, DALL-E 3 \cite{betker2023improving}, and Dreambooth \cite{ruiz2023dreambooth} have significantly improved image quality and text-image alignment by employing various approaches, including innovative training strategies and scaling of training data. Furthermore, an architectural evolution is underway, with the diffusion model framework transitioning from a U-Net structure towards a transformer-based architecture DiT ~\cite{peebles2023scalable}. PixArt-$\alpha$ \cite{chen2023pixart}, SD-3.0 \cite{esser2024scaling}, and Lumina-T2X \cite{gao2024lumina} achieve exceptional  performance through the integration of DiT. The architecture evolution blurs the previously clear delineation between diffusion and language models in the visual generative arena. In this paper, we put forward a solution based on auto-regressive generation for better quality and interactive text-guided synthesis.      

\subsection{Auto-regressive Model  for Visual Generation.}
Auto-regressive  Models~\cite{zhang2022opt,brown2020language,wei2021finetuned,touvron2023llama,vicuna2023} have been adeptly repurposed for the synthesis of visual media, including images~\cite{Muse,cogview2,Imgagen} and videos~\cite{zhang2023video,yan2021videogpt,hong2022cogvideo}. The process begins with a visual tokenizer function implemented by VQ-VAE~\cite{van2017neural} or VQ-GAN~\cite{esser2021taming}, \( f \), which effectively converts visual stimuli into a sequence of discrete tokens. Specifically, a video \( V \in \mathbb{R}^{T \times H \times W \times 3} \) (or an image when \( T = 1 \)) undergoes tokenization to yield a discrete representation \( X = f(V) \in \{1, \ldots, K\}^{T_1 \times H_1 \times W_1} \), where \( K \) denotes the codebook size intrinsic to the visual tokenizer~\cite{esser2021taming}. Subsequently, \( X \) is linearized into a one-dimensional token sequence via raster scan order, which is then introduced to a language-model transformer to facilitate generative modeling. Current auto-regressive models, include notable architectures such as ImageGPT~\cite{chen2020generative}, DALL-E~\cite{ramesh2021zero,DALLE2}, and Parti~\cite{yu2022scaling}. AR model anticipates the subsequent token based on a sequence of antecedent tokens, supplemented by additional conditional data \( c \), and adheres to a categorical distribution for \( p_{\theta}(x_i | x_{<i}; c) \). 

\begin{figure*}[!ht]
  \centering
  \includegraphics[width=\linewidth]{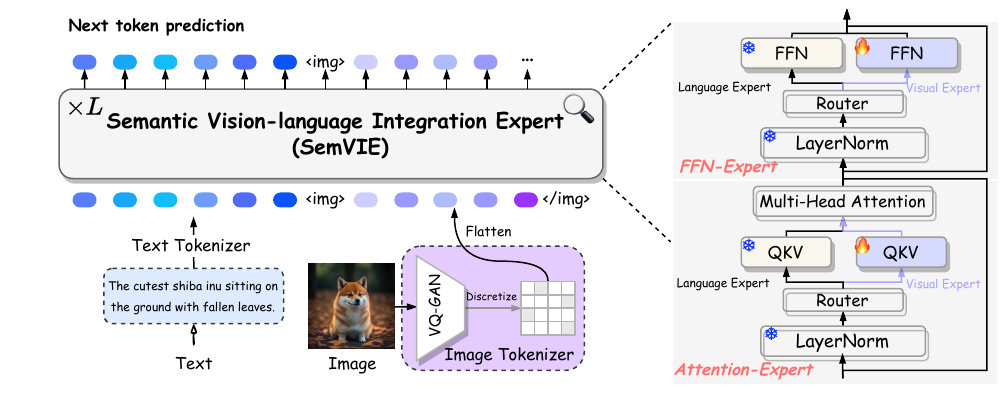} 
  \caption{Overall training framework of the proposed MARS, which consists of the SemVIE modules facilitating T2I within a unified framework.  An image-text pair is processed and tokenized by VQ-GAN \cite{esser2021taming} into 'vision words', which are then integrated with text tokens for joint processing in the SemVIE. The right part illustrates the multi-modal integration block, highlighting the synergistic processing of image and text data within the SemVIE, critical for the T2I task.}
  \label{fig:overview}
\end{figure*}
\section{Method}
\subsection{Preliminaries} \label{sec: Preliminaries}
\textbf{Auto-Regressive Models.} Auto-regressive models aim to predict future data points by regressing on their previous values. Current auto-regressive models are typically based on Transformer-like architectures \cite{vaswani2017attention}, leveraging the token prediction strategy.

\par
\noindent
\textbf{Next Token Prediction (NTP).} In the realm of sequential token analysis, one seeks to decipher the sequential arrangement, represented by the token sequence $Z = \{z_1, z_2, \ldots, z_{T_z}\}$, in which each element $z_t$ may correspond to either textual or pictorial information, encapsulated within a token, and $T_z$ denotes the sequence's aggregate length. 
The endeavor of next token prediction (NTP) is directed towards the elucidation of the auto-regressive distribution $P(z_{t+1}|z_{\leq t})$, which characterizes the likelihood of each subsequent token, thus underpinning the generative process at each juncture of the sequence. The objective of NTP is elegantly quantified through Maximum Likelihood Estimation (MLE), harnessing the negative log-likelihood, equivalently appreciated as the cross-entropy loss, articulated mathematically as:
\begin{equation}
    \mathcal{L}(\theta) = -\sum_{t=1}^{T_z - 1} \log P_\theta(z_{t+1}|z_{\leq t}),
\end{equation}
where $\theta$ symbolizes the parameters that scaffold the model, with $P_\theta(z_{t+1}|z_{\leq t})$ delineating the model's forecasted conditional probability distribution for the genesis of the ensuing token $z_{t+1}$. 

\par
\noindent
\textbf{Next K-Token Prediction (NKTP)} Next Token Prediction offers the advantages of a straightforward task format and simplicity, as well as the ability to easily extend to text-image joint generation tasks. However, when generating high-resolution images, the requirement to output long sequences results in prolonged generation times and limited image quality. To address this issue, we propose utilizing Next K Token Prediction to enhance the resolution of images generated by Next Token Prediction. Specifically, NKTP extends the NTP framework by predicting the subset of the next $K$ tokens instead of just the next single token based on predicted tokens. NKTP aims to capture longer dependencies and richer contextual information within the token sequence, enhancing the model's ability to generate coherent and contextually accurate sequences. In NKTP, the model learns to predict K tokens $\{z^i, z^{i+1}, ..., z^{i+K}\}$ given predicted tokens $\{z^j|j \leq i\}$ at each auto-regressive step:
\begin{equation}
p(z_{i+1}, z_{i+2}, \ldots, z_{i+K} | z_{\leq i})
\end{equation}

By considering multiple future tokens, NKTP can better model the dependencies between tokens, leading to more accurate and contextually appropriate predictions.

\subsection{Overall Framework}
We propose MARS, a confluence of large language models (LLMs) with vision generation capacities encapsulated within a unified framework.
MARS embodies a balanced multi-modal architecture, comprising distinct yet harmonized visual and linguistic expert models, as delineated in \cref{fig:overview}. Consistency across modalities is sustained by parallel structural designs in both modules.
The linguistic module leverages the capabilities of a pre-trained LLM, \textit{e.g.} Qwen-7B \cite{bai2023qwen}, whereas the visual counterpart undergoes initialization concomitantly with the linguistic model. During the training phase, the linguistic component remains static, and optimization is confined to select weights within the visual domain, specifically calibrated for the image synthesis task. The architecture's efficacy is further bolstered by an enriched visual vocabulary and the introduction of a \textit{SemVIE}, which amalgamates the LLM's sophisticated language interpretation abilities with visual perception. This cohesion not only harnesses the potent natural language processing capabilities inherent to the LLM but also supports the model's education across a vast corpus of paired image-text exemplars, enhancing inter-modal congruity and fostering the generation of coherent visual content.

A detailed exposition of the \textit{SemVIE} is outlined in \cref{sec: moe}. Subsequently, the manuscript explicates the nuanced process of multi-stage refinement in \cref{sec: multi-stage}. We consummate the discussion with a presentation of the meticulously curated dataset of finely annotated image-text pairings in \cref{sec: dataset construction}.

\subsection{Semantic Vision-language Integration Expert} \label{sec: moe}
\textbf{Tokenization.} In this investigation, Qwen-7B~\cite{bai2023qwen}, a pre-trained  LLM serves as the foundational linguistic framework, leveraging its tokenizer to dissect the textual data into a series of representative tokens denoted as $r_t$. Concurrently, within the visual modality, an encoder inspired by the VQ-GAN architecture \cite{lee2022autoregressive} is employed to transform the image $x \in \mathbb{R}^{3 \times H \times W}$ into a feature map $f_v \in \mathbb{R}^{K \times D}$, where, $K = H \times W / P^2$, with $P$ predefined at a quantization parameter of $16$, and $D$ encapsulates the feature dimension. The feature map $f_v$ is subsequently quantized using the visual codebook VQ-GAN that maps it onto a series of discrete code indices $f_q$. The process efficaciously refactors a $256 \times 256$ pixel image into a sequence of $256$ tokens, wherein each token embodies the information of a $16 \times 16$-pixel image segment. It is noteworthy that the visual codebook consists of $8192$ unique codices. Such visual tokens are identified within the framework as $r_v$.

In the vocab of the MARS, these visual components are interwoven with traditional textual tokens, engendering a comprehensive multimodal vocabulary. The original vocab of the linguistic LLM encompasses $151,936$ entries, which, upon symbiosis with the visual codebook and 6 special tokens(specifically designed to denote the start and end of image sequences, among other functionalities.), eventuates in a multimodal vocabulary size $160,136$. Within the architecture of MARS, visual tokens synthesized by the VQ-GAN paradigm are conferred equitable status vis-\`a-vis their textual counterparts. Initial embeddings for the visual vocab are derived from the aggregative mean embedding of pre-trained textual tokens, establishing a foundational bedrock for ensuing cross-modality integrations.

\par
\noindent
\textbf{Semantic Vision-language Integration Expert.} The MARS architecture incorporates $L$ layers of SemVIE, 
which is a specialized multi-modal Mixture of Experts (mm-MoE) designed to adeptly handle both visual 
and semantic tokens. Central to the SemVIE are the Attention-MoE and Feed-Forward Network (FFN)-MoE modules. A dedicated routing module is strategically situated following each layer normalization step within the transformer modules. This routing mechanism is designed to allocate each input token to the corresponding expert model best equipped for its processing. 
A noteworthy aspect of the shared architectural framework is the universal application of the causal multi-head attention and layer normalization modules across both language and vision modalities, epitomizing a unified methodological approach to the concurrent processing of multi-modalities data.
The process of Attention-MoE follows: 
\begin{equation} \label{eq: Attention-Moe}
\footnotesize
\begin{aligned}
\hat{r_t}, \hat{r_v} &= \text{Router}(\text{LN}(\text{Concat}(r_t, r_v)) ) \\
\hat{r_t}^{q}, \hat{r_t}^{k}, \hat{r_t}^{v} &= \frozen{W_Q^t}(\hat{r_t}), \frozen{W_K^t}(\hat{r_t}), \frozen{W_V^t}(\hat{r_t}) \\
\hat{r_v}^{q}, \hat{r_v}^{k}, \hat{r_v}^{v} &= \learn{W_Q^v}(\hat{r_v}), \learn{W_K^v}(\hat{r_v}), \learn{W_V^v}(\hat{r_v}) \\
\hat{r_q}, \hat{r_k}, \hat{r_v} &= \textbf{C}(\hat{r_t}^{q}, \hat{r_v}^{q}), \textbf{C}(\hat{r_t}^{k}, \hat{r_v}^{k}), \textbf{C}(\hat{r_t}^{v}, \hat{r_v}^{v}) \\
\hat{r} &= \text{CausualAttention} (\hat{r_q}, \hat{r_k}, \hat{r_v}) + r
\end{aligned}
\end{equation}
where \textbf{C} indicates concat operation, $\frozen{W_Q^t}$, $\frozen{W_K^t}$, and $\frozen{W_V^t}$ are frozen and loaded from pre-trained LLM. $\learn{W_Q^v}$, $\learn{W_K^v}$, and $\learn{W_V^v}$ are trainable and initialized with the pre-trained semantic LLM. Then the MoE-FFN module further processes the multi-modal tokens:
\begin{equation} \label{eq: FFN-Moe}
\small
\begin{aligned}
\hat{r_t}, \hat{r_v} &= \text{Router}(\text{LN}(\textbf{C}(r_t, r_v)) ) \\
\hat{r_t} &= \frozen{\text{FFN}^t}(r_t), \hat{r_v} = \learn{\text{FFN}^v}(r_v) \\
\hat{r} &= \textbf{C} (\hat{r_t}, \hat{r_v})
\end{aligned}
\end{equation}
where \textbf{C} indicates concat operation, $\frozen{\text{FFN}^t}$ and $\learn{\text{FFN}^v}$ share the same architecture, and $\learn{\text{FFN}^v}$ is trainable. The SemVIE module, a cornerstone of the MARS, benefits from a synergistic integration of Attention-MoE and FFN-MoE modules, enabling the effective fusion of multimodal data streams. This integration capitalizes on the profound linguistic insights afforded by the pre-trained LLM, thus leveraging the advanced language comprehension capabilities to enrich visual understanding. 
To enable the model to simultaneously predict visual tokens and text tokens, in addition to using the original LLM model head (referred to as the text head), we added a vision head to the model. \textbf{Notably,} the text token and the visual token are processed through the text head and vision head to obtain the logits, denoted as $l_t$ and $l_v$, respectively. The logits are then concatenated along the last dimension and passed through a softmax layer to obtain the probability distribution over the vocabulary for each token.


\begin{figure*}[!ht]
    \centering
    \includegraphics[width=\linewidth]{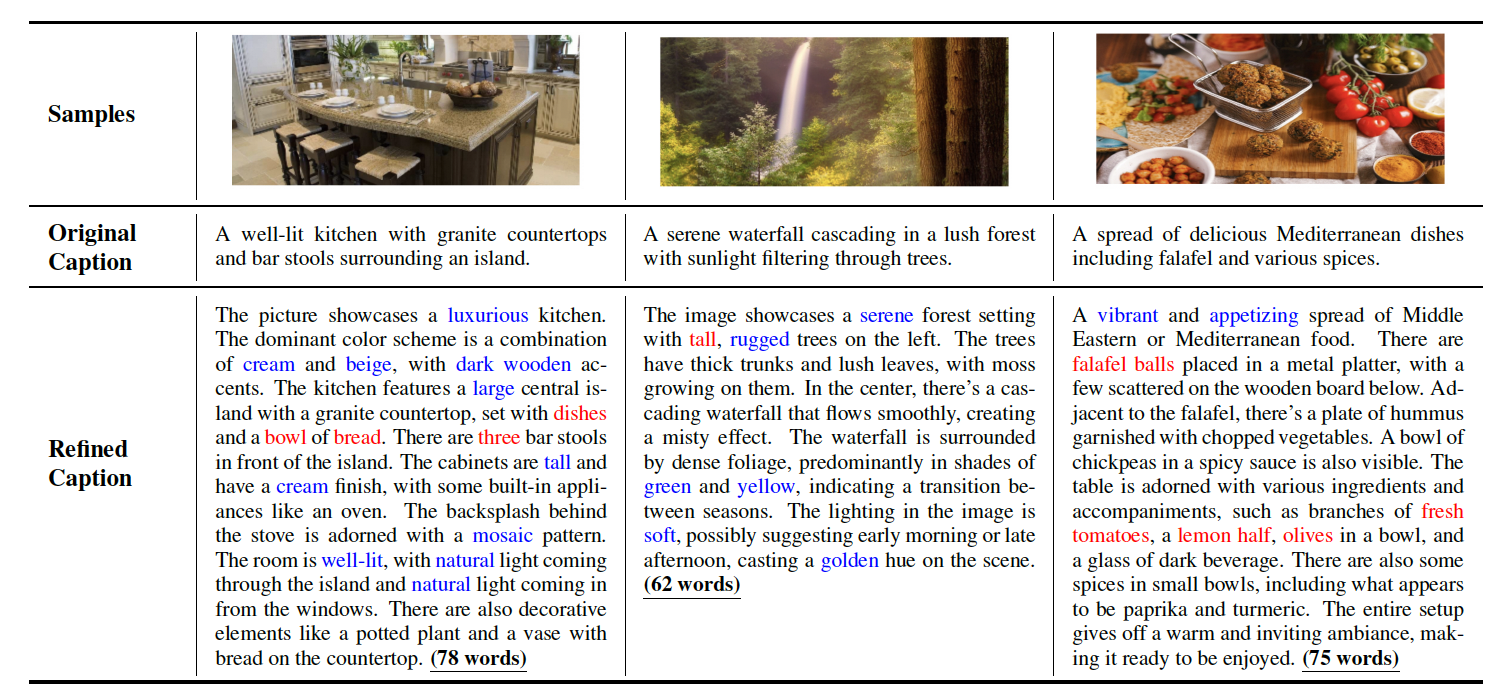}
    \caption{Comparison of dataset captions before and after reconstruction by CogVLM \cite{wang2023cogvlm}. The instruction prompt is \textit{Describe the image and its style in a very detailed manner}. The \textcolor{blue}{adjectives} are marked in blue, and \textcolor{red}{quantifiers} are marked in red to demonstrate the granularity of the reconstructed captions.}
    \label{fig:data}
\end{figure*}

\subsection{Multi-Stage Refinement} \label{sec: multi-stage}
\textbf{Stage-I: Pre-training for Text-to-Image Alignment.} We first optimize MARS by two distinct tasks: text-to-image generation and image captioning. This refinement process utilizes an auto-regressive approach for NTP, as explicated in \cref{sec: Preliminaries}. The procedure involves an extensive dataset of approximately $200$ million text-image pairs, with each image conforming to a resolution of $256 \times 256$ pixels.

\par
\noindent
\textbf{Stage-II: High-Quality Data Alignment.} To advance the fidelity of image synthesis, this stage persists in employing an NTP for the generation of images from textual descriptions. Diverging from Stage-I, the dataset enlisted for this stage comprises $50$ million pairs of text and corresponding images, each pair meticulously curated through the application of an aesthetic valuation model \cite{improvedaestheticpredictor}. The descriptive captions paired with these images originate from CogVLM \cite{wang2023cogvlm}, formulated in response to explicit directives. To mitigate potential discrepancies arising between the visual content and its textual descriptors, owing to image cropping, a standardized procedure is implemented wherein the minor axis of every image is resized to $256$ pixels. This measure, taken whilst conserving the original aspect ratio, ensures the retention of comprehensive image content. However, this results in variable sequence lengths for the images. To address this, we include resolution information in the caption to specify the desired sequence lengths of the generated images.

\par
\noindent
\textbf{Stage-III: High-Resolution Refinement.} Inspired by the approaches of SD-XL \cite{sdxl} and DeepFloyd \cite{DeepFloyd}, we utilize a cascading super-resolution strategy to further enhance MARS. The low-resolution generated images and their corresponding captions serve as inputs to the super-resolution model. The super-res model is trained after the base model has been trained. In this stage, we employ the next K-token prediction (NTKP) method to predict higher-resolution images. The output images have a long side of \textbf{1024 pixels} while maintaining the original aspect ratio. To control the resolution of the generated images, we apply the same strategy as in Stage-II. Ten million triplet (low-resolution image, caption, high-resolution image) samples were used to train the cascaded super-resolution model.

\subsection{Dataset Construction} \label{sec: dataset construction}
The open-source English datasets incorporated into our study included LAION-400M~\cite{schuhmann2021laion}, CC3M~\cite{sharma2018conceptual}, CC12M~\cite{changpinyo2021cc12m}, LAION-COCO~\cite{schuhmann2022laion}, COYO~\cite{kakaobrain2022coyo-700m}, and Datacomp~\cite{gadre2024datacomp}.
We initiate a filtration process to exclude images with resolutions below $256$ pixels or aspect ratios greater than $2$. Subsequently, we select images based on their CLIP scores ~\cite{hessel2021clipscore} and aesthetic evaluations. This methodology yields a substantial corpus of $150$ million image-text pairs. Additionally, we leveraged $50$ million in-house data, predominantly comprising image-text pairs with Chinese captions, totaling approximately $200$ million.

The coarse-grained image-text data exhibited substantial noise, evident in misalignments between images and text, deficient descriptive content, irrelevant captions, and inferior image quality. To address these challenges in the succeeding T2I instruction following the training stage, we enhance the textual relevance and informational density through a caption rewriting strategy. Specifically, we deploy a pre-trained multimodal caption model CogVLM \cite{wang2023cogvlm} to regenerate fine-grained captions for a curated selection of images. These newly generated captions intricately detail various aspects of the images, including object positioning, attributes, context, and stylistic elements, averaging approximately $110$ words in length. \cref{fig:data} showcases an illustrative sample. This approach facilitated the generation of fine-grained captions for $50$ million images.

\section{Experiment}
\label{sec:exp}

\subsection{Experiment Details}

\textbf{Implementation Details.}
We employ AdamW \cite{adamw} as the optimizer, with a beta parameter of 0.95 and weight decay set at 0.1. The peak learning rate is established at 1e-4, and a warm-up strategy is employed with a ratio of 0.01. For images with a resolution of $256 \times 256$ pixels, the batch size per GPU is set at 64, while for $512 \times 512$ pixel images, it is set at 24, leading to total batch sizes of 4096 and 1536, respectively. The training utilized DeepSpeed's ZeRO-3~\cite{rajbhandari2020zero} optimization. The training epochs for Stage-I, Stage-II, and Stage-III of the model are configured to 1, 2, and 1 epochs, respectively.

\par
\noindent
\textbf{Evaluation Benchmarks.} We select three benchmarks for comparison, including:
\begin{itemize}
    \item \textbf{MSCOCO Dataset~\cite{microsoftcoco}.} Following previous works \cite{yu2022scaling, ding2021cogview}, we generate 30k images use captions drawn from the MSCOCO 2014 evaluation dataset and assess both sample quality and image-text alignment of generated images. Specifically, we do not involve the selective curation of images from the generated output. The Fréchet Inception Distance (FID) \cite{heusel2017gans} and CLIP Score \cite{radford2021learning} are used for evaluation. 
    \item \textbf{T2I-CompBench~\cite{huang2023t2i}.} We employ various compositional prompts to assess textual attributes, including aspects such as color, shape, and texture, as well as attribute binding.
    \item \textbf{User Study.} We randomly select 100 prompts for evaluation. Subsequently, we enlist 30 participants for the user study. 
\end{itemize}

\begin{table*}[!ht] 
\centering
\caption{Quantitative evaluation of FID and CLIPScore (where available) on MS-COCO 2014 for 256 × 256 image resolution. \textbf{Diff} means diffusion model, \textbf{AR} means auto-regressive model. The results are all from the public literature. $^*$ denotes that the results are picked from the different generated images with the best CLIP score.}
\begin{tabular}{l@{\hspace{0.8cm}}|@{\hspace{0.4cm}}c@{\hspace{0.8cm}}c@{\hspace{0.8cm}}c@{\hspace{0.4cm}}|c@{\hspace{0.8cm}}c}
\toprule
\textbf{Method} &  \textbf{Venues} & \textbf{Architecture} & \textbf{\#Params} & \textbf{FID-30K $\downarrow$} & \textbf{CLIPScore $\uparrow$} \\
\midrule
GLIDE \cite{nichol2021glide}  & ICML’22 & Diff & 5.0B & 12.24 & -\\
Imagen \cite{ho2022imagen} & arXiv'22 & Diff & 3.4B & 7.27 & - \\
SDv1.0 \cite{rombach2022high} & CVPR'22 &  Diff & 1B & - & 30.50 \\
SDv1.5 \cite{rombach2022high} & CVPR'22 & Diff & -  & 9.22 & - \\
MUSE \cite{chang2023muse} & ICML'23 & Non-AR & 3B & 7.88 & \underline{32.00} 
\\
DALL-E 2 \cite{DALLE2} & arXiv'22 & Diff & 3.5B & 10.39 & 31.40 \\
PixArt-$\alpha$ \cite{chen2023pixart} & ICLR'24 & Diff & 7.32 & - \\
\midrule
DALL-E \cite{ramesh2021zero} & ICML'21 & AR & 12.0B & 28.00 & - \\
CogView \cite{ding2021cogview} & NeurIPS'21 & AR & 4.0B & 27.10 & - \\
Make-A-Scene \cite{gafni2022make}  & ECCV'22 & AR & 4.0B & 11.84 & - \\

Parti \cite{yu2022scaling} & arXiv'22 & AR & 20B & 7.23 & - \\
GILL \cite{koh2023gill} & NeurIPS'23 & AR & 6.7B & 12.20 & - \\
Emu \cite{Dai2023EmuEI} & arXiv'23 & AR & 13B & 11.70 & - \\
CM3Leon \cite{yu2023scaling}$^*$ & arXiv'23 & AR & 7B & \underline{4.88} & - \\
LAVIT \cite{jin2023unified} & ICLR'24 & AR & 7B & 7.40 & - \\
UIO-2$_{\text{XXL}}$ \cite{lu2024unified} & CVPR'24 & AR & - & 13.39 & - \\
\midrule
MARS (Ours) & - & AR & 14B & 6.92 & 32.33 \\
MARS$^*$ (Ours) & - & AR & 14B & \textbf{3.51} & \textbf{33.10} \\
\bottomrule
\end{tabular}
\label{tab:zero_shot_fid_clipscore}
\end{table*}

\begin{figure}[!ht]
    \centering
    \vspace{-0.1mm}
    \includegraphics[width=\linewidth]{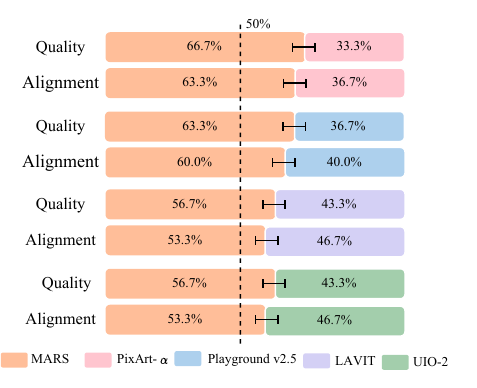}
    \caption{Human Evaluation Performance. Our MARS surpasses other state-of-the-art text-to-image models on both quality and alignment. }
    \label{fig:user-study}
\end{figure}

\subsection{Performance Comparisons and Analysis}
\textbf{MSCOCO Benchmark.} We use the Frechet Inception Distance (FID) to evaluate the quality of synthesized images. As shown in \ref{tab:zero_shot_fid_clipscore}, our proposed MARS, with only 7B trainable parameters, scores 6.92 on FID, which is a notable achievement. Compared to the auto-regressive counterpart Parti, we use fewer parameters (14B vs 20B) and smaller data sizes (0.2B vs 4.8B), achieving competitive performance (6.92 vs 7.22). Against the diffusion model SDv1.5, we achieve superior performance (6.92 vs 9.22) with less training budget (587 vs 6250 A100 GPU Days). These results highlight the efficiency of our mixture of auto-regressive models. 

Moreover, we utilize CLIP-Score to evaluate the alignment of textual conditions and corresponding generated images. MARS achieves 33.10 CLIPScore and 3.51 FID when the generated images are picked with the highest CLIP score, signaling its remarkable effectiveness in generating visually compelling imagery that closely adheres to the semantic content of the text prompts.

\begin{table*}[!ht]
    \centering
    \caption{Evaluation results (\%) on T2I-CompBench \cite{huang2023t2i}. The higher is better, and the best results are highlighted in bold.}
    \resizebox{\linewidth}{!}{
    \begin{tabular}{lccccccc}
    \toprule
    \multirow{2}{*}{Model} & \multirow{2}{*}{Venus} &   \multicolumn{3}{c}{Attribute Binding} & \multicolumn{2}{c}{Object Relationship} & \multirow{2}{*}{Complex$\uparrow$} \\
    & & Color $\uparrow$ & Shape $\uparrow$ & Texture $\uparrow$ & Spatial $\uparrow$ & Non-Spatial $\uparrow$ & \\
    \midrule
    SD1.5 \cite{rombach2022high} & CVPR'22 & 37.65 & 35.76 & 41.56 & 12.46 & 30.79 & 30.80 \\
    SDXL \cite{podell2023sdxl} & arXiv'23 & 63.69 & 54.08 & 56.37 & 20.32 & 31.10 & 40.91 \\
    Composable Diffusion \cite{liu2022compositional} & ECCV'22 & 40.63 & 32.99 & 36.45 & 8.00 & 29.80 & 28.98 \\
    Structured Diffusion \cite{feng2022training} & ICLR'22 & 49.90 & 42.18 & 49.00 & 13.86 & 31.11 & 33.55 \\
    Attn-Exct v2 \cite{chefer2023attend} & TOG'23 & 64.00 & 45.17 & 59.63 & 14.55 & 31.09 & 34.01 \\
    GORS \cite{huang2023t2i} & ICCV'23 & 66.03 & 47.85 & 62.87 & 18.15 & 31.93 & 33.28 \\ 
    DALL-E 2 \cite{ramesh2022hierarchical} & arXiv'22 & 57.50 & 54.64 & 63.74 & 12.83 & 30.43 & 36.96 \\
    PixArt-$\alpha$ \cite{chen2023pixart} & ICLR'24 & \underline{68.86} & \textbf{55.82} & \underline{70.44} & \textbf{20.82} & \underline{31.79} & \textbf{41.17} \\
    \midrule
    MARS (Ours) & - & \textbf{69.13} & \underline{54.31} & \textbf{71.23} & \underline{19.24} & \textbf{32.10} & \underline{40.49} \\
    \bottomrule
    \end{tabular}
    }
    \label{tab:t2i}
\end{table*}

\begin{table}[!t]
\centering
\caption{Ablation study of SemVIE on MS-COCO Benchmark. The term '\textit{w/o} Visual Expert' refers to a method wherein visual and text tokens are concatenated and used as inputs to fine-tune MARS without the implementation of the Visual Expert. Conversely, '\textit{w} Visual Expert' indicates the utilization of MARS's specifically designed Visual Expert architecture.}
    \begin{tabular}{l@{\hspace{-0.5cm}}|c@{\hspace{0.8cm}}c}
    \toprule
    \textbf{Method}~~~~~~~~~~ & \textbf{FID-30K $\downarrow$} & \textbf{CLIPScore $\uparrow$} \\
    \midrule
    \textit{w/o} Visual Expert~~~~~~~~~ & 10.13 & 30.14 \\
    \textit{w} Visual Expert~~~~~~~~~ & 8.24 & 31.03 \\
    \bottomrule
    \end{tabular}
\label{tab: ablation_sem}
\end{table}

\begin{table}[!t]
\centering
\caption{An ablation study assessing the impact of different stage training on the performance of the MARS model.}
    \begin{tabular}{l@{\hspace{1.7cm}}|c@{\hspace{1cm}}c}
    \toprule
    \textbf{Method} & \textbf{FID-30K $\downarrow$} & \textbf{CLIPScore $\uparrow$} \\
    \midrule
    Stage-I  & 8.24 & 31.03 \\
    Stage-II  & 7.02 & 32.21 \\
    Stage-III & 6.92  & 32.33 \\
    \bottomrule
    \end{tabular}
\label{tab:ablation_stage}
\end{table}

\par
\noindent
\textbf{T2I CompBench Performance.}
In the assessment of the T2I-CompBench, we curate a selection of contemporary text-to-image generative models for rigorous evaluation. This cohort includes Composable Diffusion \cite{liu2022compositional}, Structured Diffusion \cite{feng2022training}, Attn-Exct v2 \cite{chefer2023attend}, GORS \cite{huang2023t2i}, DALLE 2 \cite{ramesh2022hierarchical}, PixArt-$\alpha$ \cite{chen2023pixart}, SD1.5 \cite{rombach2022high}, and SD-XL \cite{podell2023sdxl}. The empirical data presented in \cref{tab:t2i} delineates the superior performance of our proposed MARS within the T2I-CompBench benchmark, underscoring its proficiency in attribute binding, delineation of object relationships, and the synthesis of intricate compositions. Notably, MARS demonstrate a marked amelioration in the fidelity of color and texture representation, achieving enhancements of +11.63\% in color fidelity and +7.49\% in texture accuracy relative to DALL-E 2. It further exhibited substantial advancements in spatial and non-spatial metrics compared to DALL-E 2, with improvements quantified at +6.41\% and +1.67\%, respectively. Moreover, when juxtaposed with the recent PixArt-$\alpha$ model, which integrates a T5-XL text encoder, MARS outperforms it in various dimensions. Specifically, MARS achieved the highest scores in color ({69.13\%}) and texture ({71.23\%}) accuracy, outperforming PixArt-$\alpha$ which scored {68.86\%} and {70.44\%} respectively. These results demonstrate that the incorporation of LLM representations and visual tokens within an auto-regressive framework can markedly improve the quality of generated images, as well as the alignment between the visual content and its corresponding textual narratives.

\par
\noindent
\textbf{User Study.}
We conduct a user study evaluating various combinations of existing methods and MARS. Each combination is assessed based on two criteria: sample quality and image-text alignment. 60 Users are asked to evaluate the aesthetic appeal and semantic accuracy of images with identical text, determining which image is superior based on these criteria. Subsequently, we calculate the percentage scores for each model, as illustrated in \cref{fig:user-study}. The results demonstrate that our MARS has significant advantages over both PixelArt-$\alpha$ and Playground-v2.5. Specifically, MARS achieves 66.7\% and 63.3\% higher voting preferences compared to PixelArt-$\alpha$ in terms of quality and alignment, respectively. Additionally, MARS shows a competitive performance when compared to LAVIT and UIO-2.

\subsection{Visual Analysis}
\begin{figure*}[!ht]
  \centering
  \includegraphics[width=\linewidth]{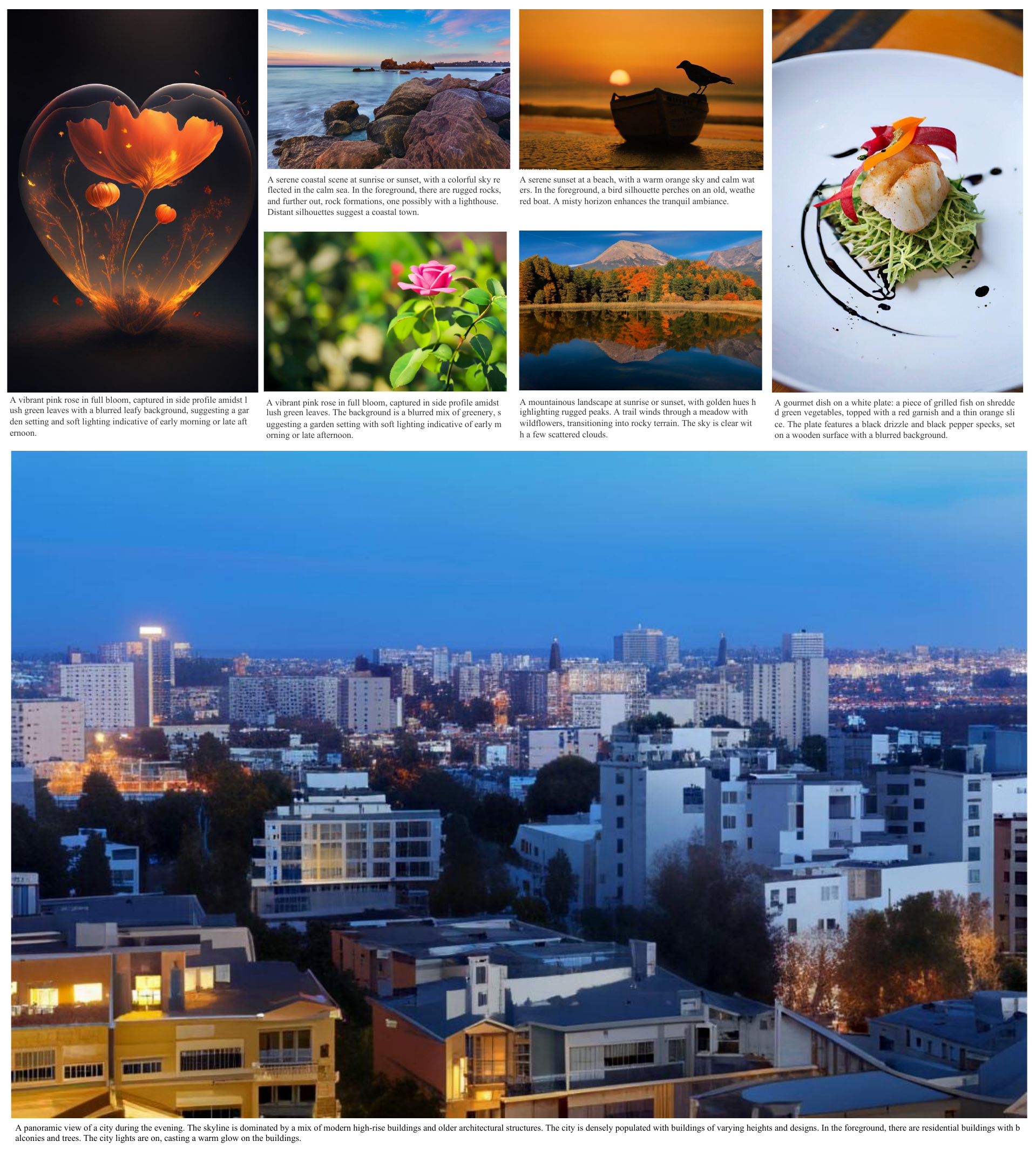}  
  \caption{Results of Visualization. The MARS framework is capable of generating realistic images across various resolutions and scenes.}
  \label{fig:more-vis}
\end{figure*}

\cref{fig:more-vis} illustrates the sophisticated image synthesis capabilities of the MARS framework, producing visuals with remarkable detail and fidelity to textual descriptions. This proficiency is likely due to the advanced textual representations extracted from Large Language Models (LLMs), which, when integrated with a structured multi-tiered training strategy, significantly improve the model's precision and alignment between text and image. The multi-stage training strategy of MARS incrementally refines the correlation between textual prompts and visual outputs, allowing for the generation of images that not only reflect the text's intent but also display a depth of detail akin to photorealistic representations. Leveraging the deep semantic understanding from LLMs, MARS adeptly translates complex textual descriptions into coherent and contextually rich visual narratives, thus exemplifying a generative model that combines technical efficiency with artistic expression.

\subsection{Multilingual Generation}
\begin{figure*}[!h]
  \centering
  \includegraphics[width=\linewidth]{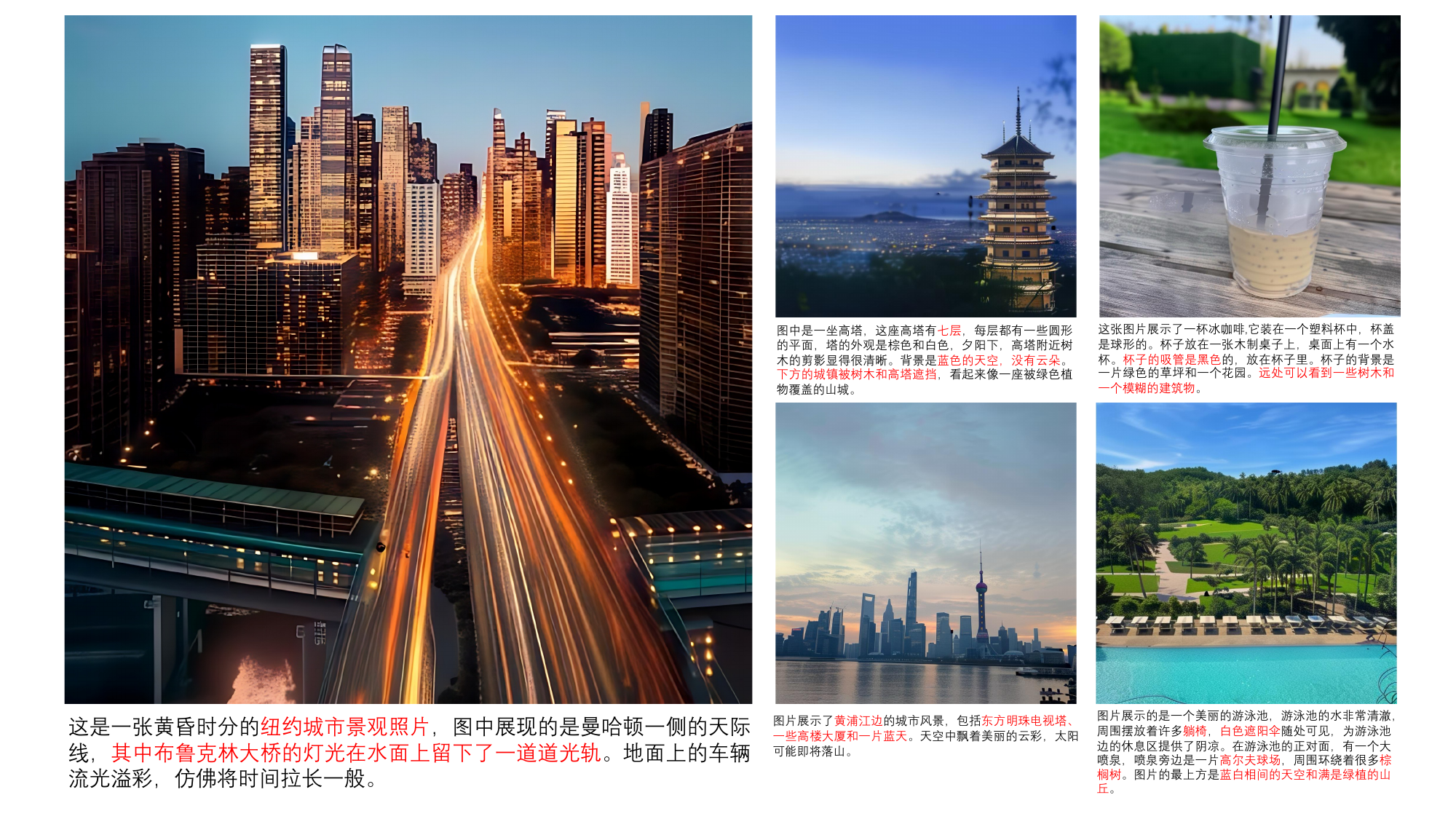}  
  \vspace{-0.5cm}
  \caption{Illustration of the model's multilingual capabilities. The model effectively responds to commands in Chinese, showcasing its inherent bilingual support and pointing toward future directions for multilingual research.}
  \label{fig:exp_fig2}
\end{figure*}

Furthermore, at the heart of our language model lies the Qwen architecture, which is intrinsically designed to support multiple languages and incorporates a comprehensive dataset featuring both Chinese and English. During the training phase, a deliberate inclusion of a small yet significant proportion of Chinese in-house data. As depicted in \cref{fig:exp_fig2}, our model attains exemplary performance in Chinese text-to-image synthesis, notwithstanding the relative scarcity of Chinese corpus. This suggests that MARS has effectively mastered the ability to interpret concepts across linguistic boundaries, ensuring that both images and text coalesce within a singular representation space, as facilitated by our novel mixture mechanism.

\begin{figure*}[!ht]
    \centering
    \includegraphics[width=\linewidth]{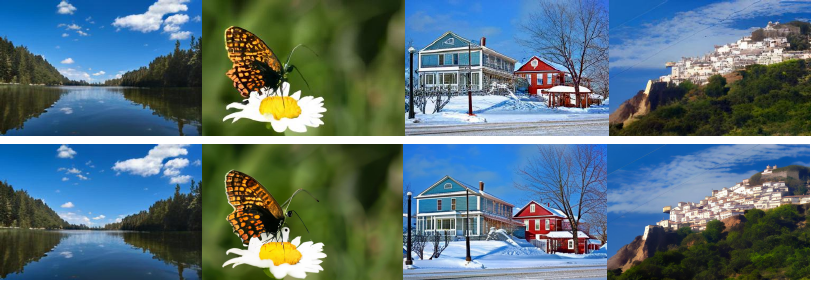}
    \caption{Qualitative comparison with different stages. The first row is from Stage II of MARS while the second row is from Stage III, indicating that Stage III is capable of correcting semantic inaccuracies while concurrently enhancing the quality of the images.}
    \label{fig:stages}
\end{figure*}

\begin{figure*}[!ht]
    \centering
    \includegraphics[width=\linewidth]{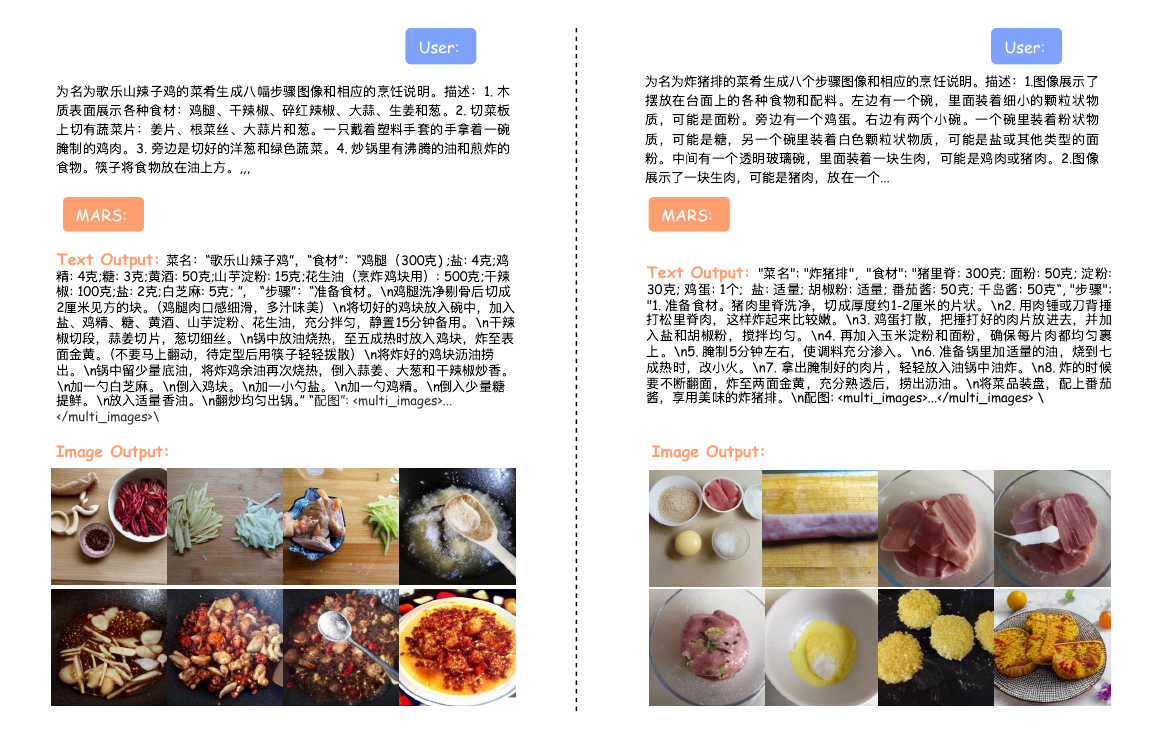}
    \caption{Example of Multimodal Recipe Generation from MARS. MARS is capable of simultaneously generating text and images. The examples of the two recipes illustrated above demonstrate that, given the recipe title and the accompanying image captions, MARS can output the recipe steps and their corresponding images in an end-to-end manner. These generated images exhibit strong relevance to the text and maintain consistency and logical coherence among themselves.}
    \label{fig: joint generation}
    \vspace{-0.5cm}
\end{figure*}

\subsection{Ablation Study}
We conduct ablation studies on the crucial parts discussed in \cref{sec: moe} and \cref{sec: multi-stage}, including model designs and multi-stage training.

\noindent
\textbf{Effect of SemVIE.} The results presented in \cref{tab: ablation_sem} were obtained during Stage-I. The \textit{w/o} Visual Expert configuration, which involves shared weights between the visual and language experts, leads to detrimental outcomes, evidenced by a 1.89 reduction in FID. This considerable decrease highlights the benefits of utilizing a specialized visual expert. The challenges associated with aligning visual and linguistic modalities underscore the need for specialized architectures that are adept at managing the intrinsic disparities between these types of data.

\noindent
\textbf{Effect of Multi-Stage.} We further explore the effect of training stages in \cref{tab:ablation_stage}. The results indicate that training in each stage positively impacts the model. On the MS-COCO benchmark, Stage II improved the FID by 1.22 compared to Stage I, and Stage III further enhanced it by 0.10 relative to Stage II. The visualizations of different stages are shown in \cref{fig:stages}. We observed that images generated during Stage I and Stage II lack detail, the images from Stage III exhibit the best quality.

\subsection{Further Analysis}
\textbf{Image and Text Joint Generation Capability.} MARS extends beyond text-to-image generation, supporting the simultaneous generation of text and images, such as generating multiple text and image outputs from text and image inputs, with a focus on the relevance, consistency, and coherence between the two modalities. Due to the preservation of LLM's integrity during MARS's pre-training phase, the system is well-positioned for tasks involving concurrent text-image creation. For instance, in the domain of recipe generation, leveraging our text-image pre-trained model, we fine-tune it with a dataset of 10,000 recipes. This enables the model to produce comprehensive cooking tutorials that include step-by-step instructions accompanied by corresponding illustrations. As depicted in \cref{fig: joint generation}, upon receiving the recipe title and associated captions requiring images, the model concurrently generates detailed textual content, such as ingredient lists and procedural steps, as well as visual representations for each stage. Notably, MARS's ability to seamlessly fuse text and imagery into coherent outputs are not confined to recipe generation and can be extrapolated to other domains requiring joint text and image generation tasks.

\section{Conclusion}
\label{sec:conclusion}
This study presents MARS, an innovative auto-regressive framework that not only retains the capabilities of pre-trained Large Language Models (LLMs) but also incorporates top-tier text-to-image (T2I) generation proficiency. MARS has been trained to exhibit exemplary performance in T2I tasks. We introduce the Semantic Vision-Language Integration Expert (SemVIE) module, which stands as the linchpin of MARS, streamlining the fusion of textual and visual token spaces and bringing a new insight into multi-modal learning. MARS has demonstrated superior performance in multiple benchmark assessments, such as the MS-COCO benchmark, T2I-CompBench, and human evaluations. 
The pre-trained Qwen model equips MARS with the ability to generate bilingual images, blending Chinese and English seamlessly. Moreover, MARS adeptly handles joint image-text generation tasks, indicating its potential for any-to-any paradigm applications. 



{
    \small
    \bibliographystyle{ieeenat_fullname}
    \bibliography{main}
}

\end{document}